\documentclass[letterpaper, 10 pt, conference]{ieeeconf/ieeeconf}

\IEEEoverridecommandlockouts                              

 \usepackage{nopageno}
\usepackage{comment}
\usepackage{amsmath}
\usepackage{svg}
\usepackage{graphicx}
\usepackage{amssymb}
\usepackage{epsfig}
\usepackage{xcolor}
\usepackage{float}
\usepackage{siunitx}
\usepackage{multirow}
\usepackage{hyperref}
\usepackage[utf8]{inputenc}
\usepackage[english]{babel}
\usepackage{makecell}
\usepackage{pifont}
\let\labelindent\relax  
\usepackage{enumitem}
\usepackage{caption}
\usepackage[noend]{algpseudocode}
\usepackage{bm}
\usepackage[hang,flushmargin]{footmisc} 
\usepackage{float}
\usepackage{changepage}
\usepackage{kantlipsum}
\newcommand{\argmaxF}{\mathop{\mathrm{argmax}}\limits}   

\usepackage{hhline}
\newlist{todolist}{itemize}{2}
\setlist[todolist]{label=$\square$}

\title{\LARGE \bf
Online Information-Aware Motion Planning with Inertial Parameter Learning for Robotic Free-Flyers
}

\author{Monica Ekal$^{1*}$, Keenan Albee$^{2*}$, Brian Coltin$^{3}$, Rodrigo Ventura$^{1}$, Richard Linares$^{2}$, and David W. Miller$^{2}$
\thanks{*Both authors contributed equally to this work.}
\thanks{$^{1}$Institute for Systems and Robotics, Instituto Superior T\'ecnico, {\tt\small\{mekal, rodrigo.ventura\}@isr.tecnico.ulisboa.pt}}
\thanks{$^{2}$ Department of Aeronautics and Astronautics, Massachusetts Institute of Technology, {\tt\small\{albee, linaresr, millerd\}@mit.edu}}
\thanks{$^{3}$ SGT Inc., NASA Ames Research Center, {\tt\small brian.coltin@nasa.gov }}
}

\begin{document}
\maketitle
\thispagestyle{plain}
\begin{abstract}
Space free-flyers like the Astrobee robots currently operating aboard the International Space Station must operate with inherent system uncertainties. Parametric uncertainties like mass and moment of inertia are especially important to quantify in these safety-critical space systems and can change in scenarios such as on-orbit cargo movement, where unknown grappled payloads significantly change the system dynamics. Cautiously learning these uncertainties en route can potentially avoid time- and fuel-consuming pure system identification maneuvers. Recognizing this, this work proposes RATTLE, an online information-aware motion planning algorithm that explicitly weights parametric model-learning coupled with real-time replanning capability that can take advantage of improved system models. The method consists of a two-tiered (global and local) planner, a low-level model predictive controller, and an online parameter estimator that produces estimates of the robot's inertial properties for more informed control and replanning on-the-fly; all levels of the planning and control feature online update-able models. Simulation results of RATTLE for the Astrobee free-flyer grappling an uncertain payload are presented alongside results of a hardware demonstration showcasing the ability to explicitly encourage model parametric learning while achieving otherwise useful motion.
\end{abstract}

\section{Introduction}
\label{sec:intro}
Robotic space systems are gearing up to perform a variety of tasks autonomously, including in-space assembly and payload transportation \cite{Brophy} \cite{flores2014review} \cite{DiFrancesco2015} \cite{Wilcox} \cite{Roque}. Precise execution of these tasks means that acceptable characterization of the dynamical system involved is often necessary. However, there is frequently underlying uncertainty in these systems; in addition to any existing model uncertainty, fuel depletion and grasping of payloads can further modify the inertial characteristics of the system during operation. Moreover, operation in cluttered, dynamic environments such as the interior of an orbiting space station calls for re-planning of trajectories in real-time to account for system and environmental changes, e.g., other free-floating payloads. Fortunately, some forms of uncertainty like inertial properties are parametric and can be resolved using knowledge of the system model. One key example of this is payload manipulation by robotic free-flyers, robots that propel themselves in microgravity. 

\begin{figure}[t!]
    \centering
    \includegraphics[width=0.80\linewidth]{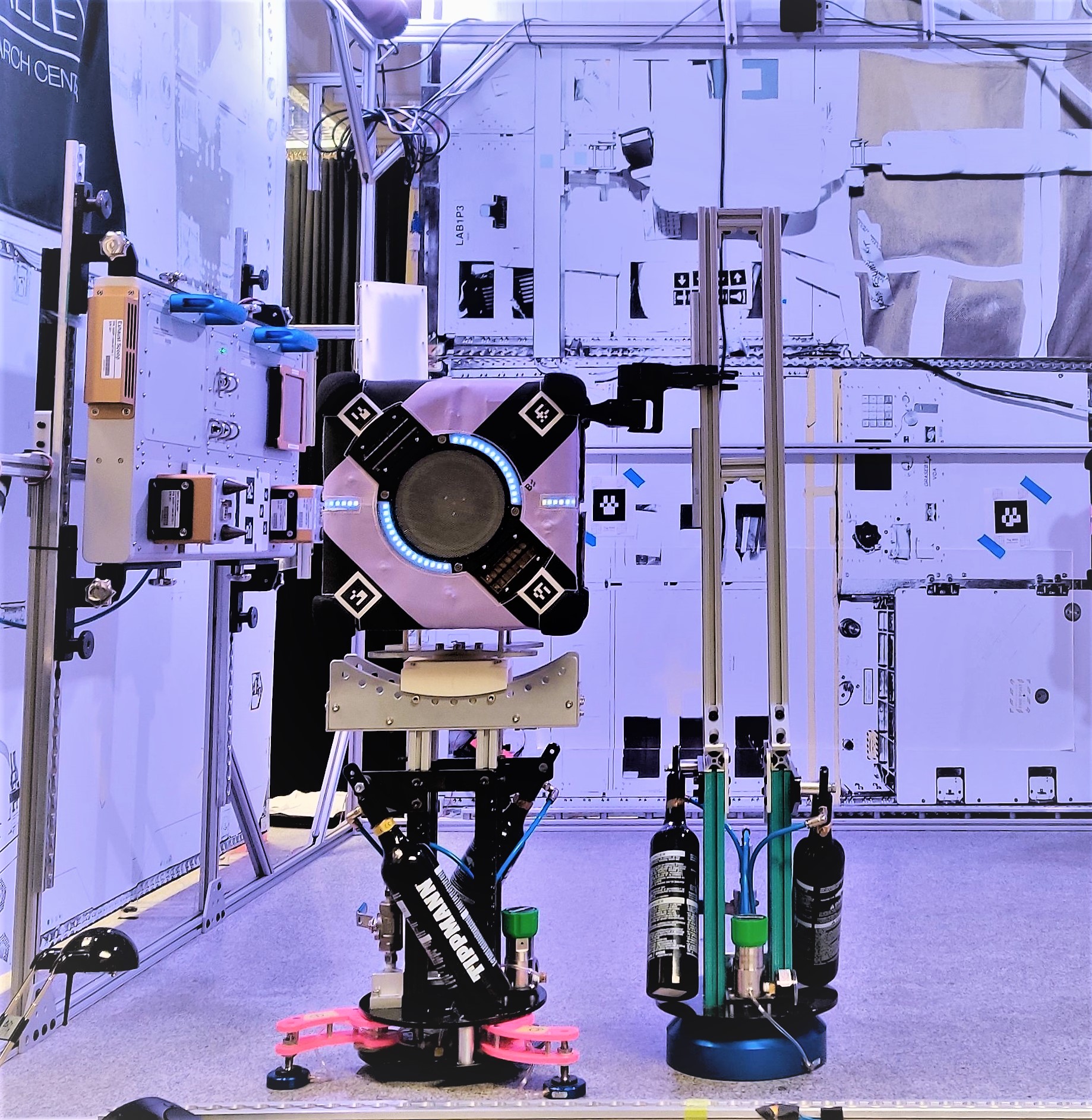}
   	\captionsetup{labelfont={bf}}
   	\caption{The Astrobee robotic free-flyer on an air-bearing grasping a new payload using its underactuated gripper.} 
   	\label{fig:Astrobee}
\end{figure}

NASA's Astrobee robot (Fig. \ref{fig:Astrobee}), recently deployed to the International Space Station (ISS), is a prime example of a free-flying robotic system \cite{fluckiger2018astrobee} \cite{SmithA} \cite{Bualat2015} \cite{coltin2016localization}. With proposed uses including automatic cargo repositioning onboard future microgravity outposts, Astrobee includes a robotic manipulator for grappling payloads and perching \cite{Park2017a}. Astrobee's proposed functionality is a key example of the need to account for system model changes and the underlying or inherited system uncertainty. Operating around moving astronauts and free-floating cargo, these systems must account for parametric model uncertainty or face poor trajectory tracking and inefficient execution, as recently shown in \cite{Albee2021}.

This places the robotic free-flying motion planning problem in the context of motion planning under parametric uncertainty. Existing planning under parametric uncertainty approaches are wide-ranging, but can be broadly placed into two categories. Some approaches attempt full system identification (sys ID) before even attempting motion planning \cite{lampariello2005modeling}, \cite{yoshida2002inertia, ekal2020accuracy} followed by planning with the estimated nominal model; otherwise, robust or chance-constrained approaches that operate under an assumed uncertainty bound \cite{How2001} \cite{majumdar2017funnel} \cite{lopez2019dynamic} are applied. Many approaches from robust and adaptive control can be applied to the uncertain tracking problem, but do not address the higher-level motion planning. Direct adaptive control \cite{Slotine} and sliding mode control can be employed to reject disturbances from parametric model uncertainty \cite{abiko2006impedance}, \cite{chen1993sliding}, \cite{ulrich2016passivity}. Indirect adaptive control approaches on the other hand, pair an estimator and a controller, relying  on  the  estimator  to  provide  a  better model  for  the  control  method under the condition of a persistently exciting control signal \cite{xu1994parameterization}, \cite{espinoza2017concurrent}. Providing robustness against assumed uncertainty bounds neglects the higher-level motion planning problem, and does not consider online reduction of uncertainty. For learnable robotic parametric unknowns, replanning capability is desirable due to an evolving understanding of the unknown parameters; further, excitation might be desired in order to aid in this model-learning process.

Relevant approaches explicitly consider the value of parametric information-awareness in the motion planning. They include a POMDP formulation  with  covariance  minimization  in  the  cost function \cite{Webb2014}, which was only demonstrated for an unconstrained double integrator system with uncertain mass, and recent work on covariance steering \cite{Okamoto2018} \cite{Okamoto2019}, which attempts to answer exactly when system excitation is most useful for model uncertainty resolution. However, these approaches have not yet been implemented on hardware, their scalability has not yet been demonstrated, and they do not address some of the practicalities of motion planning such as dealing with global long-horizon planning.

This paper proposes a method called RATTLE (\textbf{R}eal-time information-\textbf{A}ware \textbf{T}argeted \textbf{T}rajectory planning for \textbf{L}earning via \textbf{E}stimation) that combines parameter information-aware motion planning with real-time model updates. Building on the the authors' previous work \cite{albee2019combining} where the idea of making trajectory progress while accounting for information gain was explored, RATTLE proposes a user-adjustable weighting toward gaining information about parametric uncertainties to aid an online parameter estimator, with the ultimate goal of incorporating model updates. The approach is real-time receding-horizon, with the ability to also incorporate replanning. Further, the user-specified information weighting can be customized for the environment and dynamics at hand. Compared to traditional up-front sys ID, this approach only applies as much excitation as desired simultaneously with goal-achieving motion. The intent is to avoid interrupting the current maneuver and spending time on full sys ID if a sufficient amount of uncertainty reduction can instead be performed online, during otherwise useful motion. 

RATTLE is especially relevant for free-flyer load transportation scenarios, where an uncharacterized grappled payload might change the dynamics dramatically (and parametrically). Space systems requiring careful execution in cluttered space station interiors will benefit from a learning, replanning approach that is not overly conservative. To the authors' knowledge, this is the first time that a parametric information-aware planning algorithm with uncertainty reduction by parameter learning has been used for for robotic free-flyers. Though RATTLE has been employed specifically for this robotic free-flyer load transportation scenario, the algorithm's applicability extends to many systems with parametric model unknowns.

The main contributions of this paper are:
\begin{enumerate}
    \item RATTLE, a novel motion planning method for creating selectively information-aware plans with online parameter estimation to reduce parametric uncertainty;
    \item The incorporation of global motion planning into such an approach;
    \item Validation of the approach via a high-fidelity simulation of the Astrobee free-flyer transporting payload under ground testing dynamics and proof of concept results on the Astrobee hardware, demonstrating improved parametric model learning under information-aware planning.
\end{enumerate}

Section \ref{sec:intro} has introduced planning under parametric uncertainty and applications to robotic free-flyers in particular; Section \ref{sec:formulation} formulates the parametric information-aware motion planning problem and introduces the free-flying dynamics; Section \ref{sec:RATTLE} introduces RATTLE, a novel parametric information-aware motion planning algorithm; Section \ref{sec:results} demonstrates RATTLE's implementation, shows simulation and hardware results, and explains some of the method's key characteristics; Section \ref{sec:conclusion} discusses the implications of the approach and what improvements are now being pursued.

\section{Problem Formulation}
\label{sec:formulation}
A robotic system with state $\mathbf{x} \in \mathbb{R}^{n}$, input $\mathbf{u} \in \mathbb{R}^m$, and uncertain parameters $\pmb{\theta} \in \mathbb{R}^{j}$ is initially positioned at state $\mathbf{x}_0$. A region of the state space that is admissible is specified as $\mathcal{X}_{free}$, and a constraint on inputs may also be provided as $\mathcal{U}$. A goal region $\mathcal{X}_{g}$ is also specified. Let the dynamics and measurement models of the system be represented as
\begin{align}
\dot{\mathbf{x}} = f(\mathbf{x},\mathbf{u},\pmb{\theta}) + \mathbf{w}_x\\
\Tilde{\mathbf{y}} = h(\mathbf{x},\mathbf{u},\pmb{\theta}) + \mathbf{w}_y,
\label{eqn:dyn}
\end{align}
where the vector of the measured quantities is $\Tilde{\mathbf{y}} \in \mathbb{R}^{l}$, $\mathbf{w_x} \sim \mathcal{N}\left(0,\bm{\Sigma}_Q\right)$, and $\mathbf{w_y} \sim \mathcal{N}\left(0,\bm{\Sigma}_R\right)$ where $\mathcal{N}$ represents a Gaussian. Only initial estimates of the parameters are known, ${\pmb{\theta}_0} \sim \mathcal{N}(\pmb{\hat{\theta}}_0, \mathbf{\Sigma}_{\bm{\theta},0})$.

The aim is to plan a trajectory minimizing the following cost function while respecting the input and state constraints, $\mathcal{U}$ and $\mathcal{X}_{free}$,
\begin{align}
	J(\mathbf{x}, \mathbf{u}, t) = g(\mathbf{x}(t_f), \mathbf{u}(t_f)) + \int_{t_0}^{t_f}l(\mathbf{x}(t), \mathbf{u}(t))\ dt.
	\label{eqn:cont}
\end{align}

Here, $g(\mathbf{x}(t_f), \mathbf{u}(t_f))$ is a terminal cost and $l(\mathbf{x}(t), \mathbf{u}(t))$ is an accumulated cost, computed over the current nominal system model. Since knowledge of $\pmb{\theta}$ can be improved through parameter estimation, it is possible to obtain an enhanced dynamics model that is a closer representation of reality. Details on the problem setup are also provided in \cite{albee2019combining}. Even in the deterministic case, the motion planning problem is known to be at least PSPACE-hard and often requires approximate solutions \cite{Reif}.

\subsection{Rigid body dynamics}
\begin{figure}[b!]
	\centering
	\includegraphics[width=0.3\textwidth]{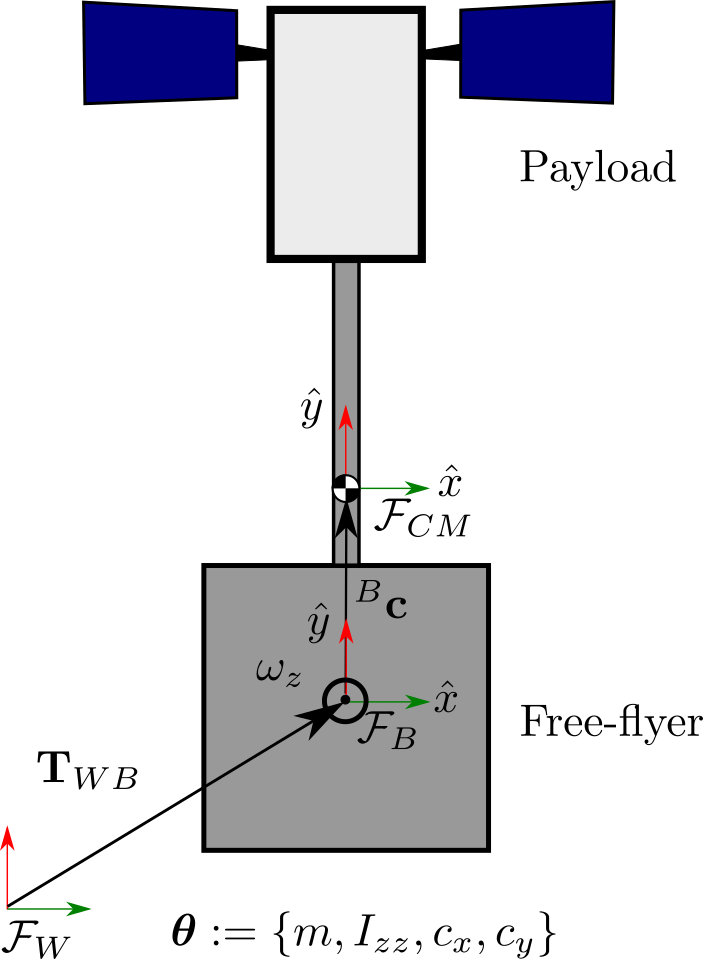} 	\captionsetup{labelfont={bf}}
	\caption{The 3 DOF rigid body dynamics model for free-flyers. An original system CM$_{0}$ located at $\mathcal{F}_B$ is offset by $\mathbf{c}$, with new system mass $m$ and $\hat{z}$ moment of inertia $I_{zz}$. Note that $\mathbf{T}_{WB}$ indicates body pose with respect to world frame.}
	\label{fig:2d}
\end{figure}

The dynamics model of interest for robotic free-flyers is the rigid body dynamics with uncertain inertial parameters. The linear and angular dynamics for a 6 DOF rigid body expressed in a body-fixed frame not coincident with the center of mass are


\begin{align}
    \begin{split}
    \begin{bmatrix}
        \mathbf{F} \\ \pmb{\tau}_{{CM}_{0}}
    \end{bmatrix}&= 
    \begin{bmatrix}
        m\mathbf{I}_3 & -m [\mathbf{c}]_{\times}\\  -m [\mathbf{c}]_{\times} & \mathbf{I}_{CM} - m[\mathbf{c}]_{\times}[\mathbf{c}]_{\times} 
    \end{bmatrix}
    \begin{bmatrix}
        \dot{\mathbf{v}} \\ \dot{\pmb{\omega}}
    \end{bmatrix} +\\
    &\begin{bmatrix}
        m [\mathbf{w}]_{\times}[\mathbf{w}]_{\times} \mathbf{c} \\
        [\mathbf{w}]_{\times} \left( \mathbf{I}_{CM} - m[\mathbf{c}]_{\times}[\mathbf{c}]_{\times} \right) \pmb{\omega}  
    \end{bmatrix}
    \end{split}
\end{align}
where ${\mathbf{v}}$, $\bm{\omega} \in \mathbb{R}^3$ denote the linear velocity and angular velocity of the original center of mass (CM$_{0}$), $\mathbf{I}_{CM}$ is the inertia tensor about the center of mass (CM), $m$ is the system mass, and $\mathbf{c} \in \mathbb{R}^3$ is the CM offset from CM$_0$. $\mathbf{F}, \bm{\tau} \in \mathbb{R}^3$ are the forces and torques applied through the $\mathcal{F}_{B}$ body frame, where $\mathcal{F}$ indicates a frame as in Fig. \ref{fig:2d}. $[-]_{\times}$ is used to indicate a cross product matrix. Note that these dynamics are significantly more complex than the Newton-Euler equations of forces and torques in the center of mass fixed frame. For a 3 DOF case commonly used in granite table free-flyer testing as in Fig. \ref{fig:Astrobee}, the equations can be written as,
\begin{gather}
F_x = m \left[ \dot{v}_{x} - \dot{\omega}_z c_y - \omega_z^2 c_x \right]\\
F_y = m \left[ \dot{v}_{y} + \dot{\omega}_z c_x - \omega_z^2 c_y \right]\\
\tau_{z_0} = m c_x \dot{v}_y - m c_y \dot{v}_x + \left[ {I_{zz,CM}} + m \left( c_y^2 + c_x^2 \right) \right] \dot{\omega}_z\\
\noindent\text{which can be conveniently grouped into matrix form,}\notag\\
\mathbf{F} = \begin{bmatrix}\mathbf{M}\end{bmatrix} \mathbf{\dot{x}} + \begin{bmatrix}\mathbf{C}\end{bmatrix}\mathbf{x}.
\end{gather}


These dynamics are also described in \cite{Jewison2014,Albee2019} and are shown for the 3 DOF case in Fig. \ref{fig:2d}. The parameter vector of interest is $\pmb{\theta} = \left\{m, c_x, c_y, I_{zz} \right\}$.

\section{Approach: Information-Aware Planning Algorithm}
\label{sec:RATTLE}
\begin{figure*}
    \centering
	\includegraphics[width=0.95\linewidth]{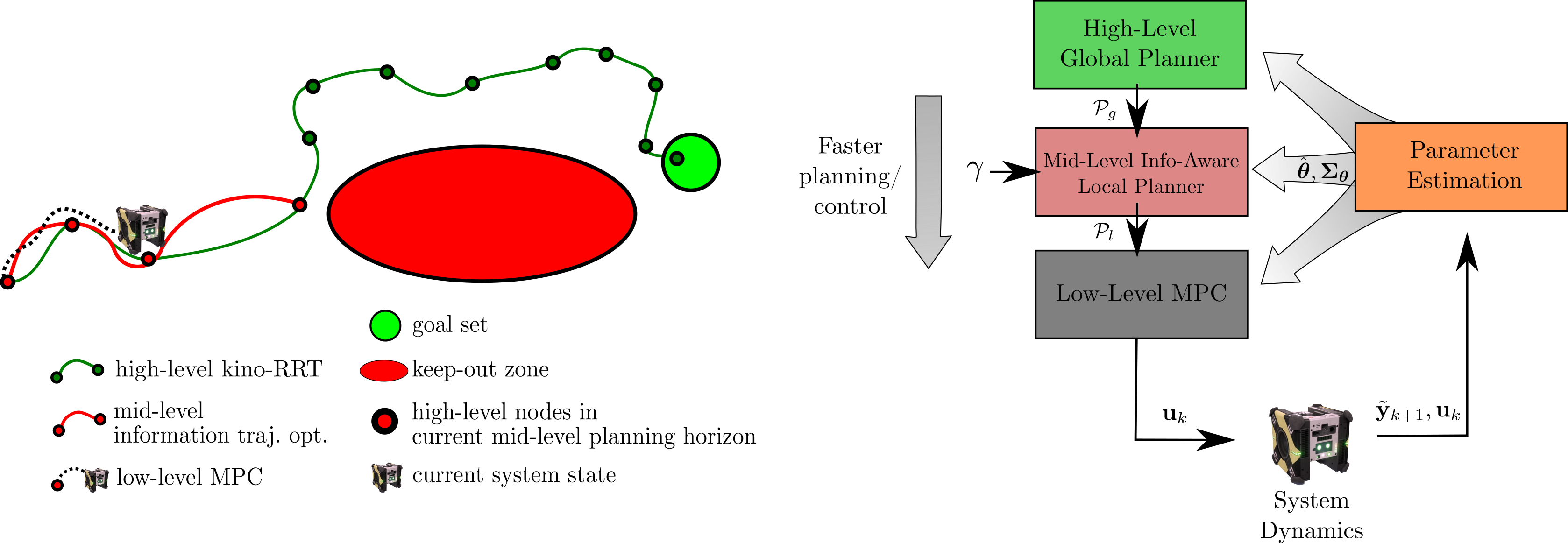}
   	\captionsetup{labelfont={bf}}
	\caption{A sketch of the RATTLE planning framework, demonstrating high-level (global) long-horizon planning via kino-RRT, and mid-level shorter-horizon (local) planning incorporating information-aware planning via an adjustable weighting term, $\gamma$. An online update-able controller, MPC, also benefits from a more accurate system model. Note that horizon lengths are not necessarily to scale.}
	\label{fig:summary}
\end{figure*}

\subsection{RATTLE Overview}

RATTLE is an information-aware motion planning method which aims to directly add informative motion when desired en route, allowing one to improve model parameter estimates online. Compared to full system identification performed prior to planning, this approach offers time savings (and potential fuel savings) by allowing useful model information to be learned en route via an explicit weighting on information-awareness in the motion planning. Compared to non-informative planning approaches, RATTLE offers a framework for trading off standard state and fuel cost minimization with the ability to perform model-improving actions; this allows the robot to take control of its level of parametric model knowledge directly via motion planning, rather than ignoring model improvement altogether.

RATTLE consists of four key ingredients:
\begin{itemize}
    \item A high-level (global) planner
    \item A mid-level (local) information-aware planner
    \item A low-level model predictive controller
    \item An online parameter estimator
\end{itemize}
As shown in Fig. \ref{fig:summary}, a global planner that excels at handling e.g., obstacle constraints and long time horizons is used to produce a nominal global plan, using a nominal set of dynamics (Section \ref{sec:global}). Portions of this global plan are used as waypoints in guiding the local planner, which incorporates an information awareness metric and operates over a shorter time horizon (Section \ref{sec:local}). Online, waypoints and information weighting may be updated at each replan of the local plan. The division into a global and local planner is in recognition of the fact that the general informative long-horizon trajectory planning problem is not computationally tractable; the common approach of using e.g., a sampling-based planner to perform global planning is proposed. At the lowest level, a model predictive controller runs at the fastest rate and continually incorporates model updates (Section \ref{sec:control}). A recursive parameter estimator runs continually, passing off the latest available model information for each planning/control element to use as desired (Section \ref{sec:est}). The RATTLE algorithm is outlined in Fig. \ref{fig:algo}. The subsections that follow describe each of these components and the estimator in further detail.

\subsection{High-Level (Global) Planner: Kinodynamic RRT}
\label{sec:global}

Sampling-based planners (SBPs) operate based on growing a tree or graph of sample points $\mathbf{x}_i$ within a sample space $\mathcal{X}_{free}$ and have been applied to a large number of robotic motion planning problems \cite{Lavalle}. A key advantage of SBPs is that difficult constraints, like collision-checking, can be explicitly checked during exploration of the state space. This framework uses kino-RRT, a variant of the popular rapidly exploring random tree (RRT) algorithm. kino-RRT includes the robot dynamics and is a good candidate for a long-horizon planner when numerical optimization-based planning becomes impractical \cite{LaValle2}. The reader is referred to Karaman for implementation specifics \cite{karaman}. Utilizing the advantage of a direct collision checking module, one may use ellipsoidal constraints for instance in order to perform simple collision checking; such constraints are common for space robotics motion planning scenarios \cite{Jewison2015}. The result of this initial long-horizon planning is a path, $\mathcal{P}_g$, of $\mathbf{x}_{0:{N_g}} \in \mathcal{X}_{free}$, where each node obeys $\mathbf{x}_{k+1} = f(\mathbf{x}_k, \mathbf{u}_k)$ and any additional enforced constraints. Dynamics propagation is typically accomplished using a set of representative $\textit{motion primitives}$. The kino-RRT is represented as the green solid line in Fig. \ref{fig:summary}, with motion primitive actions connecting adjacent waypoints. Global planning is nominally performed only once prior to motion, but online recomputation is enabled by reasonable solve times relative to the dynamics of interest (e.g., a few seconds for robotic free-flyers).

\subsection{Mid-Level (Local) Planner: Information-Aware Receding Horizon Trajectory Planner}
\label{sec:local}
The mid-level planner performs receding-horizon, information-aware planning. Starting off with the high-level, global plan $\mathcal{P}_g$ given by the kino-RRT, the planner plans trajectories between selected waypoints using updated information about the robot's model $\mathcal{M}$ based on the latest parameter knowledge $\pmb{\theta}_k$. Significantly, this planner has the ability to optimize a cost function that introduces excitation or richness in the trajectories, thus facilitating the estimation of dynamic parameters \textit{alongside} traditional state error and input use. The result is the ability to assign system excitation as desired while accomplishing otherwise useful motion.

\subsubsection{Calculation of Fisher Information}
Fisher information is employed as an information-theoretic metric in the cost. The Fisher Information Matrix (FIM) \cite{fisher1956statistical} is a measure of the amount of information given by an observation $\Tilde{y}$ about a parameter of interest, $\theta$. 
Assuming that there is no process noise in the parameter model, i.e., $\pmb{\theta}_{k+1} = \pmb{\theta}_k$, and due to the Gaussian nature of the measurement noise and linear measurement model, over time $t_0..,t_N$, the FIM is

\begin{equation}
    \mathbf{F} = \sum_{k = 0} ^{N}\mathbf{H}(t_k)^T \mathbf{\Sigma}^{-1} \mathbf{H}(t_k)
\end{equation}
\begin{equation} \begin{gathered}
    \mathbf{H}(t_k) = \frac{\partial h(\mathbf{x}(t_k),\mathbf{u}(t_k),\pmb{\theta}) }{\partial \pmb{\theta}} +\\ \frac{\partial h(\mathbf{x}(t_k),\mathbf{u}(t_k),\pmb{\theta})}{\partial \mathbf{x}}\cdot\frac{\partial \mathbf{x}(\mathbf{x}(t_k),\mathbf{u}(t_k),\pmb{\theta}) }{\partial \pmb{\theta}}.
\end{gathered} \end{equation}

More details on calculation of the FIM can be found in \cite{Wilson2014}, \cite{4046308} and the authors' previous work  \cite{albee2019combining}. A cost function is constructed to minimize the trace of the inverse of the FIM, also known as the A-optimality criterion. This is equivalent to minimizing the axis lengths of the uncertainty ellipsoid over the parameters.

As is common in trajectory optimization problems, dynamics equation (\ref{eqn:cont}) is discretized. The optimization problem solved by the mid-level planner over the horizon is then
\begin{equation}\label{opt}
\resizebox{.99\hsize}{!} {
$\begin{aligned}
    & \underset{\mathbf{u}}{\text{minimize}} & &J=
     \sum_{k = 0}^{N-1}{\mathbf{x}^T_{t+k} \mathbf{Q} \mathbf{x}_{t+k} + \mathbf{u}^T_{t+k} \mathbf{R} \mathbf{u}_{t+k}} + \gamma tr\left(\mathbf{F}^{-1}\right)\\
     & \text{subject to}
     && \mathbf{x}_{t+k+1} = f(\mathbf{x}_{t+k},\mathbf{u}_{t+k}), k = 0,..,N-1 \\
    &&& \mathbf{x}_{t+k} \in \mathcal{X}_{free}, k = 0,..,N,\\
    &&& \mathbf{u}_{t+k} \in \mathcal{U}, k = 0,..,N-1,\\
\end{aligned}$
}
\end{equation}
where $N$ is the length of the horizon and $\mathbf{Q}\succ0$ and $\mathbf{R} \succ 0$ are positive definite weighting matrices. The extent of information-richness in the trajectory can be adjusted with the relative weighting term $\gamma$. Mid-level planning occurs on timescales of approximately every few seconds, providing local plans of sufficient length to allow for system excitation without excessive recomputation ``chattering".

\subsection{Low-Level Controller: Nonlinear Model Predictive Controller}
\label{sec:control}
Model predictive control (MPC) is a control scheme that casts an optimal control problem as a mathematical optimization problem that is solved repeatedly online. Using discrete inputs as decision variables, inputs are found to minimize a cost function while satisfying constraints including the system dynamics, based on model $\mathcal{M}$. At its core MPC relies on a mathematical optimization solver to provide inputs over the designated time horizon, only the first of which is executed before recomputation is performed online. In the RATTLE framework, nonlinear MPC (NMPC) solves the optimization problem given in equation (\ref{opt}) with $\gamma = 0$ over a shorter horizon, allowing for faster control free of information metrics. NMPC was selected as the controller mainly for its ability to update model parameters on-the-fly and to incorporate input and state constraints while determining control inputs \cite{Mayne2019}. Low-level NMPC operates on timescales of approximately a few tens or hundreds of milliseconds, depending on the system of interest.

\subsection{Online Parameter Estimation: Extended Kalman Filter}
\label{sec:est}
An extended Kalman filter (EKF) is used for parameter estimation in this framework. The EKF is a non-linear extension of the Kalman filter, obtained by first-order Taylor linearization of the error dynamics about the current estimate.  Employing a filtering approach for parameter estimation in the RATTLE framework allows the estimation and thus the model updating to be performed sequentially and in real-time.
\begin{figure}[!htb]
	\begin{algorithmic}[1]
	\Procedure{\texttt{RATTLE}}{$\mathbf{x}_0, \mathcal{X}_g, \pmb{\theta}_0, \mathbf{Q}, \mathbf{R}, \mathcal{M}$}
	\State \texttt{InitParamEst($\pmb{\theta}_0$)}
	\State $\mathcal{P}_g \gets$ \texttt{GlobalPlan($\mathbf{x}_0, \mathcal{X}_g, \mathcal{M}$)}
	\State $k=0$
	\While {$\mathbf{x}_k \not\in \mathcal{X}_g $}
		\If{$\mathcal{P}_g \texttt{ replan requested}$}
			\State $\mathcal{P}_g \gets$ \texttt{GlobalPlan($\mathbf{x}_k,  \pmb{\theta}_k, \mathcal{X}_g, \mathcal{M}$)}
		\EndIf
		\State $\gamma \gets \texttt{GetInfoWeight}(k)$
		\State $\mathcal{P}_l, N_l \gets \texttt{LocalInfoPlan}(\mathbf{x}_k, \pmb{\theta}_k, \mathcal{P}_g, \mathcal{M}, \gamma)$
		\While{$k < N_{l}$}
			\State $\mathbf{u}_{k} \gets \texttt{NmpcControl}(\mathbf{x}_k, \pmb{\theta}_k, \mathcal{P}_l, \mathcal{M})$
			\State $\mathbf{x}_{k+1} = f(\mathbf{x}_k, \mathbf{u}_{k})$  \Comment{system dynamics}
			\State $\pmb{\theta}_{k+1} \gets \texttt{ParamEst} (\pmb{\theta}_k, \mathbf{\tilde{y}}_{k}, \mathbf{u}_k)$
			\State $k = k + 1$
		\EndWhile
	\EndWhile
	\EndProcedure
	\State
	
	\Procedure{\texttt{LocalInfoPlan}}{$\mathbf{x}_k, \pmb{\theta}_k, \mathcal{P}_g, \mathcal{M}, \gamma$}
	\State $\mathcal{M} \gets \texttt{UpdateModel}(\pmb{\theta}_k, \mathcal{M})$
	\State \texttt{UpdateCost}$(k)$
	\State $\mathbf{x}_{N}, N_l \gets \texttt{UpdateWaypoint}(k, \mathcal{P}_g)$
	\State $\mathcal{P}_l \gets \texttt{RunTrajOpt}(\mathbf{x}_k, \mathbf{x}_N, \texttt{CalcFisher}(), \mathcal{M})$
	\State \Return{$\mathcal{P}_l, N_l$}
	\EndProcedure
	\end{algorithmic}
	\caption{The algorithmic overview of the RATTLE framework. Note that $\mathcal{M}$ indicates a system model, consiting of $f(-)$ and $h(-)$ as in equation \ref{eqn:dyn} with accompanying constraints $\mathcal{X}_{free}$ and $\mathcal{U}$. $N_l$ indicates a local plan horizon length index, and $\mathcal{P}_{[-]}$ indicates a plan, i.e., a set of $\mathbf{x}_{k:k+N}$ and $\mathbf{u}_{k:k+N}$ over a time horizon.}
	\label{fig:algo}
\end{figure}
\section{Results}
\label{sec:results}
RATTLE was validated in a high-fidelity simulator of the free-flyer dynamics of NASA's Astrobee robot; a proof of concept demonstration of the information-aware planning and parameter estimator was also carried out on Astrobee hardware at the NASA Ames granite table facility\footnote{A granite table is a near-frictionless surface used for simulating the microgravity free-flyer dynamics. They also require impeccable cleaning for successful tests, which the authors were able to partake in firsthand.}. 

The Astrobee free-flyer is a cube-shaped robot, measuring 32 cm per side \cite{Bualat2015}. Its holonomic propulsion system draws in air through two central impellers, which is expelled precisely through 12 exhaust nozzles for thrust \cite{SmithA}. The Astrobee Robot Software uses ROS as middleware for communication, with about 46 nodelets grouped into approximately 14 processes running on two ARM processors \cite{fluckiger2018astrobee} \cite{fluckiger2018astrobee2}. The Astrobee Robot Software consists of a simulator, which enables testing of developed algorithms before implementation on hardware. The simulator is essentially a set of plug-ins for the Gazebo robot simulator, which offer the same ROS interfaces as the hardware. The ROS/Gazebo-based simulation environment includes extensive modeling of Astrobee including its impeller propulsion system, onboard visual navigation, environmental disturbances, and many more true-to-life models \cite{fluckiger2018astrobee}.

A few key properties of the motion planning method were demonstrated. Primarily, the ability of the method to selectively add parameter information gathering was shown by setting informative values of $\gamma$. The convergence of system parameter estimates was then compared to tests in which no information weighting was provided. This illustrated the improved quality of parameter estimates with on-the-fly parameter learning, thus offering the ability to make goal-achieving plans that also accomplish parameter learning, as opposed to conventional system identification. The full RATTLE pipeline was demonstrated in simulation to show the selective addition of informativeness to goal-achieving plans. Hardware results were also obtained specifically for the mid-level planner, showing targeted parameter learning for a regulation task. Ground truth parameter values used for both experiments are shown in Table \ref{tab:params}.

\begin{figure}[hbt!]
    \centering
	\includegraphics[width=0.85\linewidth]{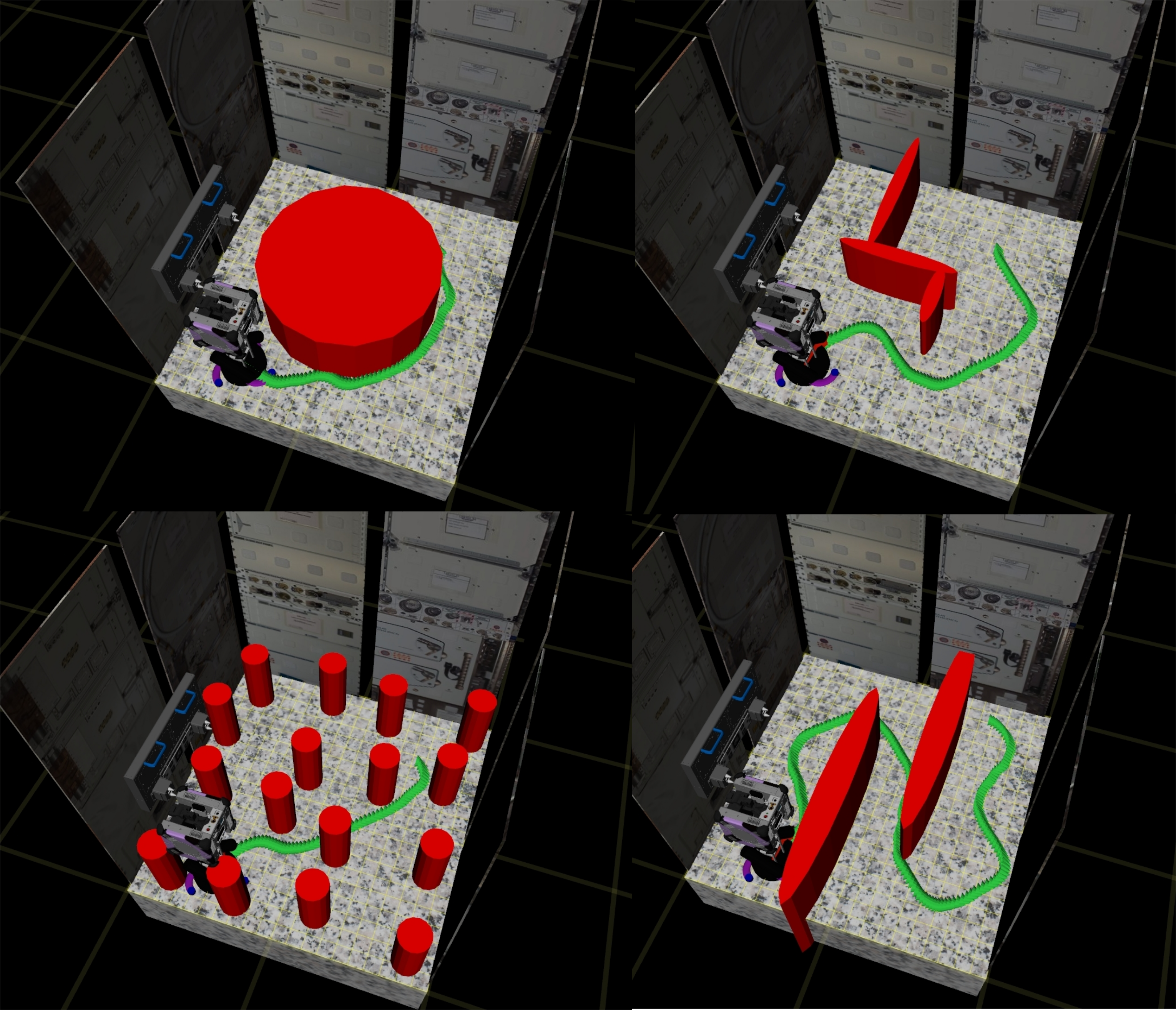}
   	\captionsetup{labelfont={bf}}
	\caption{Examples of the global planner under different obstacle constraints. Here, global plans are shown in green with obstacles shown in red.}
	\label{fig:rrt}
\end{figure}

\subsection{Simulation Demonstration: RATTLE in the Astrobee High-Fidelity 3 DOF Simulation}
\begin{figure*}[bt!]
  	\centering
    \includegraphics[width=1.0\linewidth]{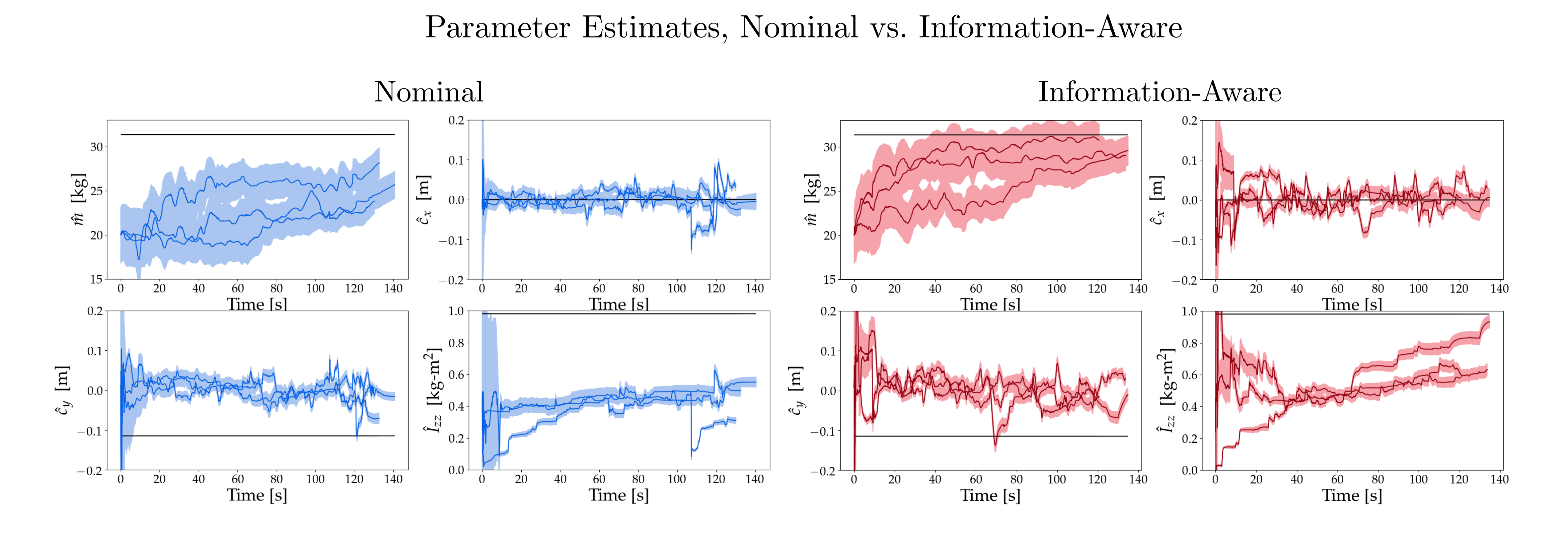}
  	\captionsetup{justification=centering,labelfont={bf}}
  	\caption{Parameter estimates while tracking without information-aware planning (blue), and with information-awareness (red) for the robot-grasping-payload system in simulation. Four parameters of interest are shown for each case, and ground truth values are shown in black. Three runs of each case are illustrated with 1-sigma confidence shown as a highlight.}
  	\label{fig:params}
  		\vspace{-1.5em}	
\end{figure*} 

\begin{figure}[hbt!]
    \centering
	\includegraphics[width=0.65\linewidth]{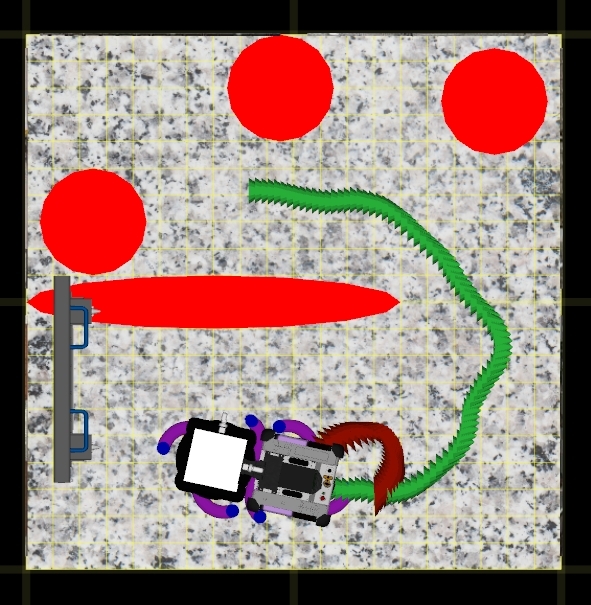}
   	\captionsetup{labelfont={bf}}
	\caption{Top-down view of the test setup used for Fig. \ref{fig:params}, representing a room with a narrow opening and cluttered obstacles inside (inflated for Astrobee radius). Here, the global plan can be seen in green with a local plan (with some information weighting) in red.}
	\label{fig:excite}
\end{figure}

The RATTLE framework was implemented in the high-fidelity Astrobee simulation\footnote{https://github.com/nasa/astrobee} to demonstrate its capabilities for a 3 DOF cargo re-positioning scenario, matching the environment and dynamics used for hardware testing.  After Astrobee rigidly grapples a payload as in Fig. \ref{fig:Astrobee} or Fig. \ref{fig:excite}, the ground truth parameters $\pmb{\theta} = \left\{m, c_x, c_y, I_{zz} \right\}$ change and parametric uncertainty enters the problem. Equipped with the payload, Astrobee was tasked with moving within a tolerance of a goal region $\mathcal{X}_g$. Note that Astrobee has severe input limits of $u_{max} \leq 0.4\ [N]$, meaning that system inertial parameters are particularly important to know before safety critical maneuvering is needed.

A kinodynamic RRT for the translational states was used as a global planner. Some examples of the planner's flexibility are shown in Fig. \ref{fig:rrt}, which also shows an example of the granite table simulation environment. Any real-time global planning method could be used, but kinodynamic RRT was selected because it uses dynamics model knowledge in its planning but with real-time computation capabilities. An $n=50$ Monte Carlo set of test runs was performed on the randomly reordered obstacle world of the bottom left of Fig. \ref{fig:rrt} to demonstrate real-time properties. Running on a quad-core Intel i7-4700MQ machine alongside the full Astrobee autonomy stack, a C++ implementation of the global planner computed plans in $3.59 \pm 3.63\ [s]$ for $\sim2\ [m]$ global plans with obstacle density of $\sim30\%$. This was a particularly challenging scenario---in practice for tests as in Fig. \ref{fig:excite} runtimes were usually below $0.5\ $ [s].

\begin{figure}[hbt!]
   	\centering
    \includegraphics[width=1.0\linewidth]{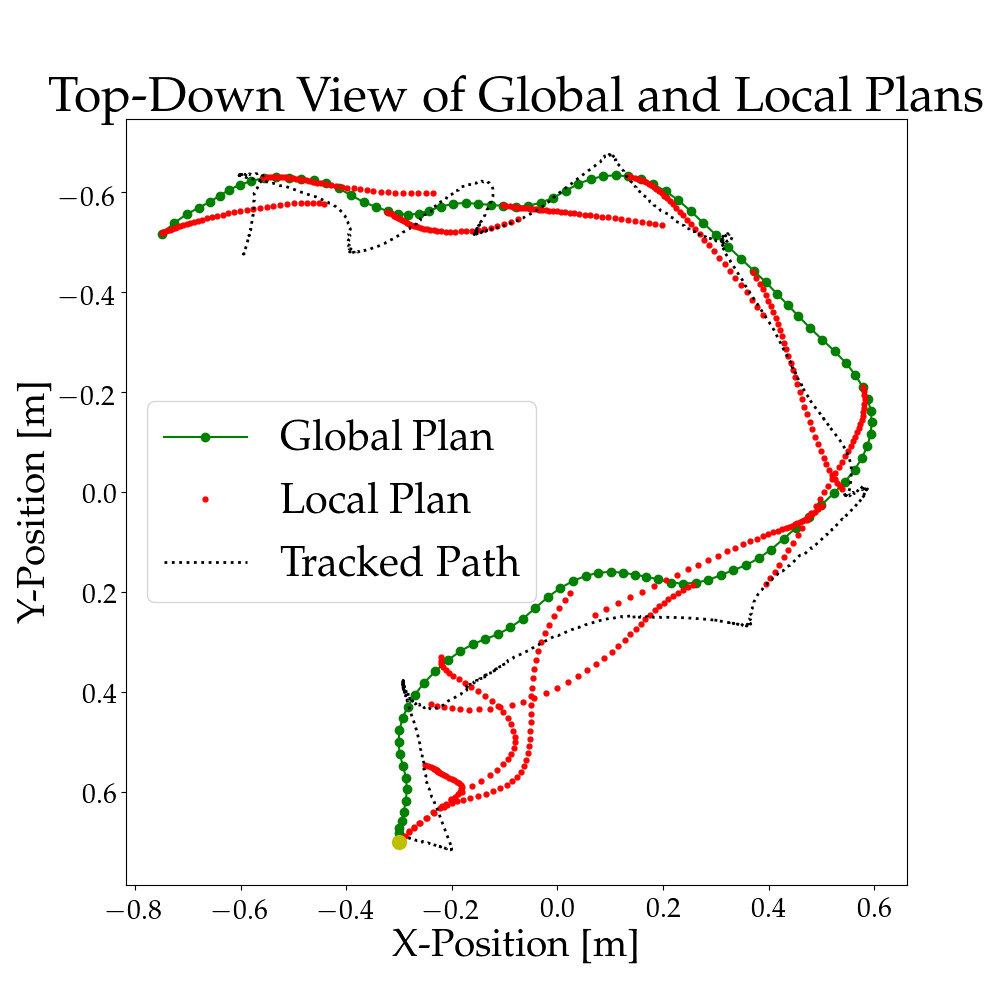}
   	\captionsetup{labelfont={bf}}
   	\captionsetup{justification=centering,labelfont={bf}}
   	\caption{An example of a robot-grasping-payload information-aware trajectory using RATTLE, in simulation. The yellow dot denotes the start point. Note the reduction of excitation in local plans towards the end of the trajectory.}
   	\label{fig:rattle_demo}
   		\vspace{-1.5em}	
   \end{figure}

The ACADO toolkit \cite{Houska2011a} was used to implement the nonlinear programming-based information-aware local planner running with a replan period of $12\ $[s] and the low-level model predictive controller running at $10\ $[Hz]. The information-aware mid-level planning scheme of Section \ref{sec:local} was used, with an exponentially decaying weighting on $\mathbf{\gamma}$ with time constant $\tau$ of $\frac{1}{10}$ the global plan horizon. The number of local replans used was 11. A sample run of RATTLE can be seen in Fig. \ref{fig:rattle_demo}, where the global plan is tracked by local plans containing desired levels of information-awareness. As this weighting decreases and estimation accuracy improves, the controller and planner models resemble the system behaviour to a greater degree of accuracy. The parameter estimator used poses and twists from Astrobee's localization algorithm, along with the applied forces and torques as inputs.  Estimated parameters were incorporated into the system model of the local planner and controller at a period of $16\ $[s]. This avoided updates using transient estimates and controller instabilities due to a rapidly changing system model. The parameter estimation comparison is shown explicitly in Fig. \ref{fig:params}, where non-informative plans for three representative runs are shown at left (blue) compared to information-aware plans at right (red). $\hat{m}$ and $\hat{I}_{zz}$ in particular show improvement in information-aware plans, while poor observability rendered accurate center of mass estimates difficult to obtain for both cases.

\begin{table}[hbt!]
	\centering
	\begin{tabular}{c|c c c}
		& \thead{Astrobee+\\Arm+Carriage} &\thead{Payload+\\Carriage} & \thead{Combined System\\($I_{zz}$ about CM)} \\
		\hline
		\text{Sim}\\
		$m\ $[kg]      & 19.568 & 11.8 & 31.368\\
		$I_{zz}\ $[kg-m$^2$] & 0.282 & 0.015   & 0.980\\
		$c_{x}\ $[m] & 0.0 & 0.0   & 0.0\\
		$c_{y}\ $[m] & 0.0 & -.305  & -0.115\\
		\hhline{====}
		\text{Hardware}\\
		$m\ $[kg]      & $19.0$ & 11.8 & 30.8\\
		$I_{zz}\ $[kg-m$^2$] & $0.25$ & 0.015   & 0.94\\
		$c_{x}\ $[m] & $0.0$ & $0.0$   & 0.0\\
		$c_{y}\ $[m] &0.0  & -.305   & -0.12\\
		
	\end{tabular}
	\captionsetup{labelfont={bf}}
	\caption{Simulation and hardware ground truth values. Note that hardware values are approximations, accounting for gas level, arm extension, and number of batteries used.}
	\label{tab:params}
\end{table}

\subsection{Hardware Demonstration: Information-Aware Motion Planning Proof of Concept on 3 DOF Astrobee Testbed}
A series of hardware tests were conducted on the Astrobee free-flyer granite table facility, using a ``without payload" and ``with payload" configuration, shown in Fig. \ref{fig:Astrobee}, using nominal and information-aware versions of the mid-level planner. Experiments included a three-waypoint maneuver; variance of estimates of the targeted parameters was compared post-maneuver between nominal and information-aware planning, with results indicated in Table \ref{tab:hardware}. The mid-level planner ran onboard Astrobee, providing real-time updates at 3 Hz. $I_{zz}$ in particular saw a large reduction in variance when information-aware planning was used, as rotational excitation was not as frequently used in nominal planning. This indicates the dramatic affect of intentional excitation in parameter information-awareness for parameters which are not otherwise excited; notably, mass saw little variation between nominal and information-aware planning since nominal plans already include translational excitation.

\begin{table}[hbt]
	\centering
	\begin{tabular}{|c|c|c|}
		\hline
		& Without Payload & With Payload\\
		\hline
		\thead{$I_{zz}$ Covariance \\ $[$\% Change$]$}     & -25.01\% &  -38.05\%\\
		\hline
		\thead{$m$ Covariance \\ $[$\% Change$]$}  & 2.47\% &  -3.71\%\\
		\hline
	\end{tabular}
	\captionsetup{labelfont={bf}}
	\caption{Parameter estimate variance reduction of information-aware plans relative to non-informative plans for hardware testing. Decreases indicate greater precision of estimated model parameters. Both ``without payload" and ``with payload" cases are shown (average of three runs at the final timestep of motion).}
	\label{tab:hardware}
\end{table}

\section{Conclusion}
\label{sec:conclusion}
This paper introduced RATTLE (Real-time information-Aware Targeted Trajectory planning for Learning via Estimation) for robotic systems operating under parametric uncertainty. Particularly relevant for free-flyer cargo transportation scenarios, this method encourages model-learning through information-awareness in motion planning while fulfilling the primary control objectives. A sampling-based global planner (kinodynamic RRT) and a receding horizon planner that maximizes information content of a local trajectory constitute the planning module. Non-linear model predictive control (NMPC) is used for trajectory tracking. An online filtering estimator, EKF in this case, provides real-time model updates for all planning and control elements, providing an improved system model for future use. The ability of this framework to plan for information gain and the resulting improvement in estimate accuracy was validated with results from high-fidelity 3 DOF simulation of the Astrobee free-flyer, as well as granite table hardware testing using Astrobee; video results are available\footnote{https://youtu.be/Kim32sjs2VM}. Future work aims to expand robustness guarantees to the approach, refine methods of updating the global plan, discuss RATTLE tuning in further detail, and explore the application of the approach to 6 DOF Astrobee cargo transportation on the International Space Station. 

\section*{Acknowledgments}
Funding for this work was provided by the NASA Space Technology Mission Directorate through a NASA Space Technology Research Fellowship under grant 80NSSC17K0077. This work was also supported by the LARSyS - FCT Plurianual funding 2020-2023, P2020 INFANTE project 10/SI/2016, and an MIT Seed Project under the MIT Portugal Program. The authors gratefully acknowledge the support that enabled this research. The authors would like to thank Marina Moreira, Ruben Garcia Ruiz, and the Astrobee team at NASA Ames for their help in setting up the hardware for experiments. Thank you to Alex Cabrales and Oliver Jia-Richards for insightful conversations.
\bibliographystyle{ieeetr}
\bibliography{main}

\begin{thebibliography}{10}

\bibitem{Brophy}
J.~R. Brophy, L.~Friedman, N.~J. Strange, D.~Landau, T.~Jones, R.~Schweickart,
  C.~Lewicki, M.~Elvis, and D.~Manzella, ``{Synergies of Robotic Asteroid
  Redirection Technologies and Human Space Exploration},'' in {\em 65th
  International Astronautical Congress}, 2014.

\bibitem{flores2014review}
A.~Flores-Abad, O.~Ma, K.~Pham, and S.~Ulrich, ``A review of space robotics
  technologies for on-orbit servicing,'' {\em Progress in Aerospace Sciences},
  vol.~68, pp.~1--26, 2014.

\bibitem{DiFrancesco2015}
J.~M. DiFrancesco and J.~M. Olson, ``{The economics of microgravity
  research},'' {\em Nature Partner Journals - Microgravity}, vol.~1, nov 2015.

\bibitem{Wilcox}
B.~H. Wilcox, ``{ATHLETE: A Cargo-Handling Vehicle for Solar System
  Exploration},'' in {\em IEEE Aerospace Conference}, 2011.

\bibitem{Roque}
P.~Roque and R.~Ventura, ``{Space CoBot: modular design of an holonomic aerial
  robot for indoor microgravity environments},'' tech. rep.

\bibitem{fluckiger2018astrobee}
L.~Fluckiger, K.~Browne, B.~Coltin, J.~Fusco, T.~Morse, and A.~Symington,
  ``Astrobee robot software: A modern software system for space,'' 2018.

\bibitem{SmithA}
T.~Smith, J.~Barlow, M.~Bualat, T.~Fong, C.~Provencher, H.~Sanchez, and
  E.~Smith, ``{Astrobee: A New Platform for Free-Flying Robotics on the ISS},''
  in {\em iSAIRAS 2016}, 2016.

\bibitem{Bualat2015}
M.~Bualat, J.~Barlow, T.~Fong, C.~Provencher, and T.~Smith, ``{Astrobee:
  Developing a Free-flying Robot for the International Space Station},'' in
  {\em AIAA SciTech}, 2015.

\bibitem{coltin2016localization}
B.~Coltin, J.~Fusco, Z.~Moratto, O.~Alexandrov, and R.~Nakamura, ``Localization
  from visual landmarks on a free-flying robot,'' in {\em Proc. of the Int.
  Conf. on Intelligent Robots and Systems (IROS)}, pp.~4377--4382, 2016.

\bibitem{Park2017a}
I.~W. Park, T.~Smith, H.~Sanchez, S.~W. Wong, P.~Piacenza, and M.~Ciocarlie,
  ``{Developing a 3-DOF compliant perching arm for a free-flying robot on the
  International Space Station},'' {\em IEEE/ASME International Conference on
  Advanced Intelligent Mechatronics, AIM}, pp.~1135--1141, 2017.

\bibitem{Albee2021}
K.~Albee and A.~C. Hernandez, ``{The Case for Parameter-Aware Control of
  Assistive Free-Flyers},'' in {\em AIAA SciTech GNC}, 2021.

\bibitem{lampariello2005modeling}
R.~Lampariello and G.~Hirzinger, ``Modeling and experimental design for the
  on-orbit inertial parameter identification of free-flying space robots,'' in
  {\em ASME 2005 International Design Engineering Technical Conferences and
  Computers and Information in Engineering Conference}, pp.~881--890, American
  Society of Mechanical Engineers Digital Collection, 2005.

\bibitem{yoshida2002inertia}
K.~Yoshida and S.~Abiko, ``Inertia parameter identification for a free-flying
  space robot,'' in {\em AIAA Guidance, Navigation, and Control Conference and
  Exhibit}, p.~4568, 2002.

\bibitem{ekal2020accuracy}
M.~Ekal and R.~Ventura, ``On the accuracy of inertial parameter estimation of a
  free-flying robot while grasping an object,'' {\em Journal of Intelligent \&
  Robotic Systems}, vol.~98, no.~1, pp.~153--163, 2020.

\bibitem{How2001}
J.~P. How and M.~Tillerson, ``{Analysis of the impact of sensor noise on
  formation flying control},'' {\em Proceedings of the American Control
  Conference}, vol.~5, pp.~3986--3991, 2001.

\bibitem{majumdar2017funnel}
A.~Majumdar and R.~Tedrake, ``Funnel libraries for real-time robust feedback
  motion planning,'' {\em The International Journal of Robotics Research},
  vol.~36, no.~8, pp.~947--982, 2017.

\bibitem{lopez2019dynamic}
B.~T. Lopez, J.~P. Howl, and J.-J.~E. Slotine, ``Dynamic tube mpc for nonlinear
  systems,'' in {\em 2019 American Control Conference (ACC)}, pp.~1655--1662,
  IEEE, 2019.

\bibitem{Slotine}
J.-J.~E. Slotine and W.~Li, ``{Applied Nonlinear Control},''

\bibitem{abiko2006impedance}
S.~Abiko, R.~Lampariello, and G.~Hirzinger, ``Impedance control for a
  free-floating robot in the grasping of a tumbling target with parameter
  uncertainty,'' in {\em 2006 IEEE/RSJ International Conference on Intelligent
  Robots and Systems}, pp.~1020--1025, IEEE, 2006.

\bibitem{chen1993sliding}
Y.-P. Chen and S.-C. Lo, ``Sliding-mode controller design for spacecraft
  attitude tracking maneuvers,'' {\em IEEE transactions on aerospace and
  electronic systems}, vol.~29, no.~4, pp.~1328--1333, 1993.

\bibitem{ulrich2016passivity}
S.~Ulrich, A.~Saenz-Otero, and I.~Barkana, ``Passivity-based adaptive control
  of robotic spacecraft for proximity operations under uncertainties,'' {\em
  Journal of Guidance, Control, and Dynamics}, vol.~39, no.~6, pp.~1444--1453,
  2016.

\bibitem{xu1994parameterization}
Y.~Xu, H.-Y. Shum, T.~Kanade, and J.-J. Lee, ``Parameterization and adaptive
  control of space robot systems,'' {\em IEEE transactions on Aerospace and
  Electronic Systems}, vol.~30, no.~2, pp.~435--451, 1994.

\bibitem{espinoza2017concurrent}
A.~T. Espinoza and D.~Roascio, ``Concurrent adaptive control and parameter
  estimation through composite adaptation using model reference adaptive
  control/kalman filter methods,'' in {\em 2017 IEEE Conference on Control
  Technology and Applications (CCTA)}, pp.~662--667, IEEE, 2017.

\bibitem{Webb2014}
D.~J. Webb, K.~L. Crandall, and J.~{Van Den Berg}, ``{Online parameter
  estimation via real-time replanning of continuous Gaussian POMDPs},'' {\em
  Proceedings - IEEE International Conference on Robotics and Automation},
  pp.~5998--6005, 2014.

\bibitem{Okamoto2018}
K.~Okamoto, M.~Goldshtein, and P.~Tsiotras, ``{Optimal Covariance Control for
  Stochastic Systems under Chance Constraints},'' {\em IEEE Control Systems
  Letters}, vol.~2, no.~2, pp.~266--271, 2018.

\bibitem{Okamoto2019}
K.~Okamoto and P.~Tsiotras, ``{Optimal Stochastic Vehicle Path Planning Using
  Covariance Steering},'' {\em IEEE Robotics and Automation Letters}, vol.~4,
  no.~3, pp.~2276--2281, 2019.

\bibitem{albee2019combining}
K.~Albee, M.~Ekal, R.~Ventura, and R.~Linares, ``Combining parameter
  identification and trajectory optimization: Real-time planning for
  information gain,'' {\em arXiv preprint arXiv:1906.02758}, 2019.

\bibitem{Reif}
J.~Reif, ``{Complexity of the Generalized Mover's Problem},'' tech. rep.,
  Harvard University, 1985.

\bibitem{Jewison2014}
C.~M. Jewison, D.~Miller, and A.~Saenz-Otero, ``{Reconfigurable Thruster
  Selection Algorithms for Aggregative Spacecraft Systems},'' 2014.

\bibitem{Albee2019}
K.~Albee, {\em {Toward Optimal Motion Planning for Dynamic Robots: Applications
  On-Orbit}}.
\newblock Master's thesis, Massachusetts Institute of Technology, 2019.

\bibitem{Lavalle}
S.~M. LaValle, {\em {Planning Algorithms}}, vol.~9780521862.
\newblock 2006.

\bibitem{LaValle2}
S.~LaValle and J.~Kuffner, ``{Randomized Kinodynamic Planning},'' {\em The
  International Journal of Robotics Research}, 2001.

\bibitem{karaman}
S.~Karaman and E.~Frazzoli, ``{Sampling-based Algorithms for Optimal Motion
  Planning},'' {\em The International Journal of Robotics Research}, 2011.

\bibitem{Jewison2015}
C.~Jewison, R.~S. Erwin, and A.~Saenz-Otero, ``{Model Predictive Control with
  ellipsoid obstacle constraints for spacecraft rendezvous},'' {\em
  IFAC-PapersOnLine}, vol.~28, no.~9, pp.~257--262, 2015.

\bibitem{fisher1956statistical}
R.~A. Fisher, ``Statistical methods and scientific inference.,'' 1956.

\bibitem{Wilson2014}
A.~D. Wilson, J.~A. Schultz, and T.~D. Murphey, ``{Trajectory synthesis for
  fisher information maximization},'' {\em IEEE Transactions on Robotics},
  vol.~30, no.~6, pp.~1358--1370, 2014.

\bibitem{4046308}
J.~H. {Taylor}, ``The cramer-rao estimation error lower bound computation for
  deterministic nonlinear systems,'' in {\em 1978 IEEE Conference on Decision
  and Control including the 17th Symposium on Adaptive Processes},
  pp.~1178--1181, Jan 1978.

\bibitem{Mayne2019}
J.~B. Rawlings, D.~Q. Mayne, and M.~M. Diehl, {\em {Model predictive control:
  theory, computation, and design}}, vol.~197.
\newblock 2019.

\bibitem{fluckiger2018astrobee2}
L.~Fluckiger, K.~Browne, B.~Coltin, J.~Fusco, T.~Morse, and A.~Symington,
  ``Astrobee robot software: Enabling mobile autonomy on the iss,'' in {\em
  Proc. of the Int. Symposium on Artificial Intelligence, Robotics and
  Automation in Space (i-SAIRAS)}, 2018.

\bibitem{Houska2011a}
B.~Houska, H.~Ferreau, and M.~Diehl, ``{ACADO} {T}oolkit -- {A}n {O}pen
  {S}ource {F}ramework for {A}utomatic {C}ontrol and {D}ynamic
  {O}ptimization,'' {\em Optimal Control Applications and Methods}, vol.~32,
  no.~3, pp.~298--312, 2011.

\end{thebibliography}
\end{document}